\documentclass[letterpaper, 10 pt, conference]{ieeeconf}  

\IEEEoverridecommandlockouts                              

\usepackage{graphicx} 
\usepackage{epsfig} 
\usepackage{amsmath} 
\usepackage{amssymb}  
\usepackage{booktabs}
\usepackage{xcolor} 
\usepackage{cite}
\usepackage{url}
\usepackage{caption}
\usepackage{multirow}
\usepackage{flafter}
\usepackage{arydshln} 

\title{\LARGE \bf
Retrieve-to-Localize: Bridging Large Language Models and LiDAR Geometry for Spatial Grounding}

\author{Byounggun Park$^{1,*}$, Giyong Moon$^{2,*}$, Jusung Kim$^{1}$, and Soonmin Hwang$^{2,\dagger}$
\thanks{$^{*}$Equal contribution.}%
\thanks{$^{1}$Department of Automotive Engineering (Automotive-Computer Convergence), Hanyang University, Seoul, South Korea.}%
\thanks{$^{2}$Department of Automotive Engineering, Hanyang University, Seoul, South Korea.}%
\thanks{$^{\dagger}$Corresponding author: soonminh@hanyang.ac.kr}%
}

\begin{document}

\maketitle
\thispagestyle{empty}
\pagestyle{empty}

\begin{abstract} 
LiDAR provides precise geometric information for spatial perception tasks such as object detection in autonomous driving and outdoor robotics. However, recognizing and localizing individual objects is not sufficient to answer questions that require composing spatial relations and grounding the intended target. Motivated by recent advances in large language models (LLMs) for autonomous driving, we leverage their language priors to interpret complex spatial questions and ground the referred target in LiDAR geometry. To support this spatial grounding capability, we introduce SpatialLiDAR-QA, which combines single- and multi-step relational grounding with complementary spatial understanding tasks. We further propose SpatialLiDAR-LM, which aligns LiDAR point features with an LLM and grounds target coordinates through language-conditioned, position-aware proposal retrieval and local point refinement. This design derives target coordinates directly from local LiDAR geometry rather than through textual language decoding. Experiments demonstrate substantial improvements over representative LiDAR--language models and multi-camera VLMs on precise coordinate prediction tasks. Our dataset and model training code will be publicly released.
\end{abstract}

\section{INTRODUCTION}
\label{sec:introduction}

\begin{figure}[t]
\centering
\includegraphics[width=\linewidth]{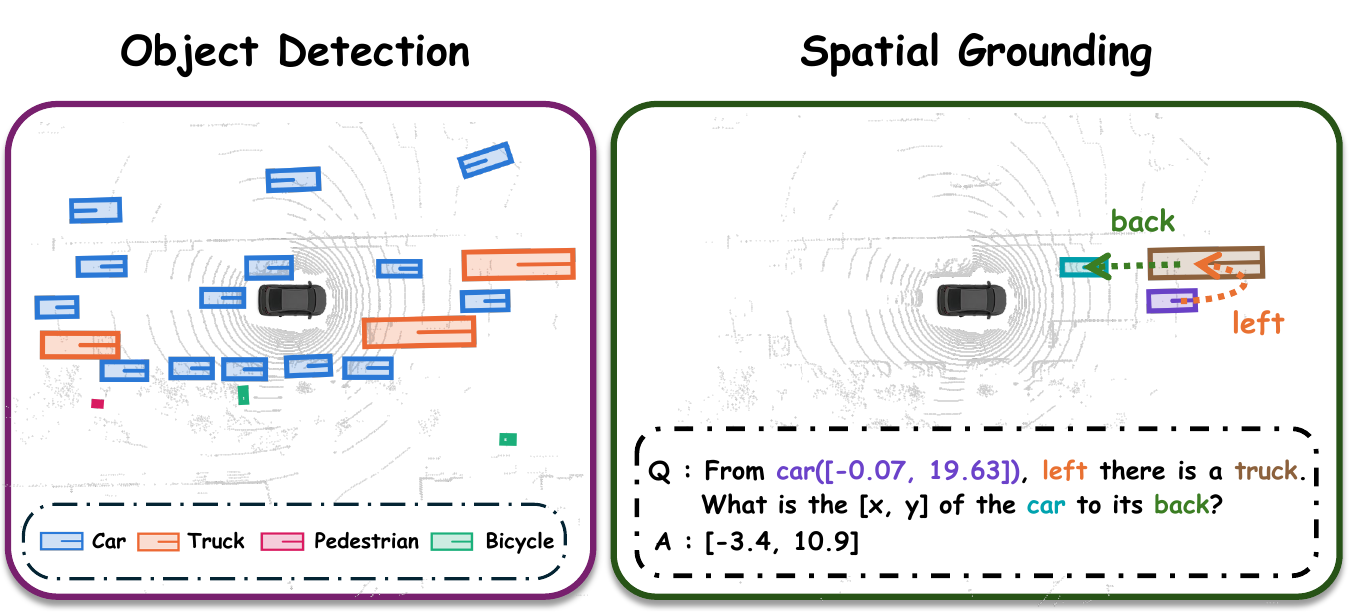}

\caption{LiDAR provides precise geometry for object detection, but spatial questions require more than localizing individual objects: relations must be composed to ground the intended target.}
\label{fig:teaser}
\vspace{-6mm}
\end{figure}

Recent advances in multimodal learning have enabled language-based understanding across diverse modalities, including images~\cite{liu2023llava,bai2023qwenvl}, speech and audio~\cite{tang2024salmonn}, point clouds~\cite{hong20233dllm,xu2024pointllm}, and radar~\cite{guan2025talk2radar}. These developments have encouraged the use of large language models (LLMs) in autonomous driving, connecting sensor observations with question answering, reasoning, and decision-making~\cite{sima2024drivelm,tian2025drivevlm}. In driving and outdoor robotics environments, such capabilities require understanding not only what objects are present, but also where they are and how they are spatially related.

Within this broader research direction, LiDAR--language models have established capabilities in object detection, captioning, and question answering~\cite{yang2024lidarllm,choi2025b4dl}. Building on these advances, we focus on spatial grounding: questions that couple relational reasoning with precise localization. For example, a question may ask for the object in front of another object that is itself to the left of a specified reference. Answering it requires identifying the reference, following the first relation to an intermediate object, and resolving the second relation to the final target. Predicting the target's class and coordinates then requires connecting this relational interpretation to the observed geometry. As illustrated in Fig.~\ref{fig:teaser}, this motivates tasks that jointly require reference identification, relation composition, and precise coordinate prediction, rather than evaluating object recognition or location prediction in isolation.

Constructing such tasks requires spatial relations that are clearly specified and supervision that connects their interpretation to target locations. Existing datasets provide relational understanding, object detection, and geometry-based spatial evaluation~\cite{qian2024nuscenesqa,yang2024lidarllm,tian2025nuscenesspatialqa}. In particular, some relational QA formulations represent object-to-object directions through six angular sectors referenced to the ego vehicle's forward axis~\cite{qian2024nuscenesqa}. These sectors support relational questions but describe broad spatial regions with limited metric specificity. Their boundaries can also make directional labels sensitive to small changes in relative position. When multiple objects satisfy the same directional relation, the relation label alone may not uniquely identify the intended referent. Categorical relational answers alone do not verify whether the target is precisely localized, while object detection alone fails to test whether the target is identified through relational reasoning. Supervision that systematically connects these capabilities for both training and evaluation remains limited. These gaps motivate a dataset that combines clearly defined relations with single- and multi-step relational grounding requiring precise coordinate predictions, together with complementary spatial understanding tasks to address these challenges.

To this end, we introduce \textbf{SpatialLiDAR-QA}, a LiDAR-native dataset of 108,811 QA pairs constructed from nuScenes keyframes~\cite{caesar2020nuscenes}. We define direct directional relations from relative object-center coordinates, with the specified reference object determining the relational origin and the ego vehicle defining fixed directional axes. The generation process filters out directionally ambiguous cases and composes valid relations into single- and multi-step relational grounding questions that require predicting the final target's class and LiDAR-frame coordinates. Complementary Localization and Object Identification tasks assess object-to-coordinate and coordinate-to-class correspondence, while maneuver feasibility evaluates the use of surrounding geometry for ego-centric decisions. Together, these tasks provide spatial grounding supervision, but a model must still distinguish the referred instance from similar candidates and recover its coordinates from the supporting LiDAR geometry. Existing LiDAR--language models demonstrate that aligned scene features can support spatial grounding. Nevertheless, decoding location tokens from these representations does not explicitly derive the predicted coordinates from the points supporting the selected target~\cite{yang2024lidarllm}.

We therefore propose \textbf{SpatialLiDAR-LM}, an end-to-end architecture that aligns point features with an LLM. Its Point-Retrieved Localization (PRL) module uses the LLM's contextual language representation to retrieve question-relevant spatial proposals and refine the target coordinate from their supporting local points. Proposal-Key Positional Encoding incorporates proposal locations into retrieval, helping distinguish similar instances according to the spatial conditions expressed in the question. This design connects language-conditioned target selection with explicit spatial geometry, enabling the model to use LiDAR's spatial information when answering spatial grounding questions.

Our contributions are summarized as follows:
\begin{itemize}
    \item We introduce \textbf{SpatialLiDAR-QA}, which couples explicitly defined spatial relations with single- and multi-step relational grounding and complementary spatial understanding tasks, supporting both training and evaluation of LiDAR--language spatial grounding.
    \item We propose \textbf{SpatialLiDAR-LM}, which connects contextual language representations to language-conditioned, position-aware proposal retrieval and local point refinement, deriving target coordinates from question-relevant LiDAR geometry.
    \item We evaluate SpatialLiDAR-LM on SpatialLiDAR-QA and demonstrate substantial improvements over representative LiDAR--language and multi-camera models on precise coordinate prediction.
\end{itemize}

\section{RELATED WORK}
\label{sec:related_work}

\subsection{LiDAR-Based MLLMs for Outdoor Scene Understanding}

Recent LiDAR-based MLLMs have extended language understanding to outdoor point-cloud scenes. BEV-LLM~\cite{11097781} combines LiDAR and multi-view image features in a unified BEV representation for scene captioning and uses absolute positional encoding to generate view-specific descriptions. LiDAR-LLM~\cite{yang2024lidarllm} supports captioning, question answering, and 3D grounding by aligning detector-pretrained BEV features with an LLM. Its Position-Aware Transformer divides BEV features into six camera-aligned views and injects the corresponding positional embeddings into the features and queries, and its grounding task generates location tokens specifying 3D boxes. B4DL~\cite{choi2025b4dl} extends LiDAR--language modeling to sequential frames for spatio-temporal tasks, including temporal reasoning and time grounding, using aligned frame-wise features and ego-motion metatokens.

Building on these advances, our model connects relational spatial reasoning with precise coordinate estimation from LiDAR geometry.

\subsection{Spatial Reasoning Datasets for Autonomous Driving}

Existing autonomous-driving QA datasets provide useful supervision for scene understanding, but differ in how spatial relations are represented and evaluated. NuScenes-QA~\cite{qian2024nuscenesqa} constructs 3D scene graphs with object-to-object relations defined by six angular categories relative to the ego vehicle's forward axis. Although this supports relational questions, each category describes a broad spatial sector, and its existence, counting, recognition, status, and comparison tasks produce categorical answers rather than precise target coordinates. Its formulation therefore evaluates relational QA without directly requiring precise localization of the referred object. LiDAR-LLM~\cite{yang2024lidarllm} provides captioning and 3D grounding supervision, but its captioning tasks are generated from camera images using 2D MLLMs and GPT-4 filtering, and its tasks do not systematically combine precise localization with compositional object-relative reasoning. NuScenes-SpatialQA~\cite{tian2025nuscenesspatialqa} uses ground-truth 3D geometry to construct spatial understanding and reasoning questions. However, its object references are derived from camera-image captions, and it is designed for image-based VLM evaluation rather than providing training supervision for LiDAR--language alignment. To complement these benchmarks, SpatialLiDAR-QA supports both training and evaluation on tasks requiring spatial reasoning, such as single- and multi-step relational grounding with precise coordinate prediction.

\begin{figure*}[t]
\centering
\includegraphics[width=\linewidth]{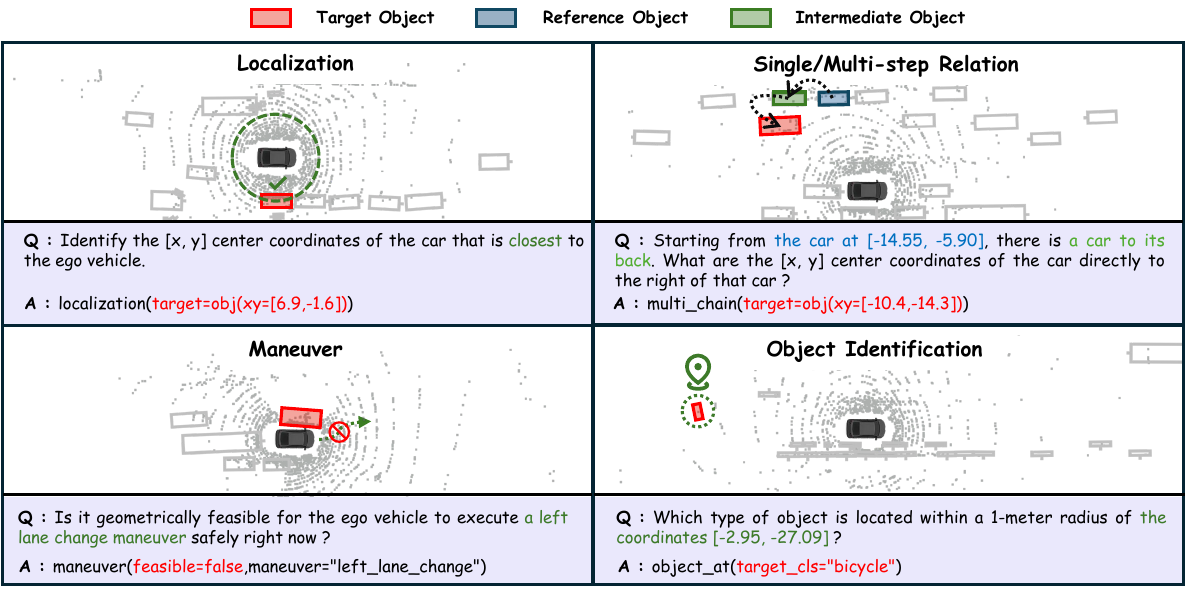}
\caption{Overview of the five spatial grounding tasks in SpatialLiDAR-QA:
Direct Localization, Object Identification, Single/Multi-Step Relational Grounding,
and Maneuver Feasibility, all formulated from LiDAR observations.}
\label{fig:task}
\vspace{-4mm}
\end{figure*}

\section{SpatialLiDAR-QA Dataset}
\label{sec:dataset}

As illustrated in Fig.~\ref{fig:task}, SpatialLiDAR-QA defines five complementary tasks for training and evaluation.

\subsection{Dataset Details}
\label{subsec:dataset_details}

\textbf{1. Direct Localization:} 
Given an object category, the model predicts the LiDAR-frame center coordinates of the instance \textit{closest} to the ego vehicle, testing distance-based instance selection and precise localization.

\begin{table}[t]
\centering
\caption{Distribution of QA pairs in SpatialLiDAR-QA.}
\label{tab:dataset_stats}
\resizebox{\columnwidth}{!}{%
\begin{tabular}{lrrr}
\toprule
\textbf{Task Category} & \textbf{Training} & \textbf{Validation} & \textbf{Total} \\
\midrule
Localization & 24,586 & 5,273 & 29,859 \\
Object Identification & 5,998 & 1,303 & 7,301 \\
Single-step Relational & 27,050 & 5,826 & 32,876 \\
Multi-step Relational & 8,152 & 2,400 & 10,552 \\
Maneuver Feasibility & 23,762 & 4,461 & 28,223 \\
\midrule
\textbf{Total} & \textbf{89,548} & \textbf{19,263} & \textbf{108,811} \\
\bottomrule
\end{tabular}%
}
\end{table}

\textbf{2. Object Identification:} 
Given a LiDAR-frame coordinate $(x, y)$, the model predicts the object class at that location, testing spatial-to-semantic correspondence.

\textbf{3. Single-Step Relational Grounding:} 
Given a reference object and a directional relation along the ego-aligned axes, the model predicts the target's class and LiDAR-frame coordinates. Partially occluded targets are also included, subject to the criteria in Sec.~\ref{subsec:dataset_construction}.

\textbf{4. Multi-Step Relational Grounding:} 
The model composes two directional relations through an intermediate object to predict the final target's class and LiDAR-frame coordinates, keeping the ego-aligned axes fixed.

\textbf{5. Maneuver Feasibility:} 
Given the surrounding object configuration, the model judges whether continuing straight or changing lanes to either side is geometrically feasible, testing ego-centric free-space reasoning.

\textbf{Dataset Statistics.} Table~\ref{tab:dataset_stats} summarizes the distribution of QA pairs across task categories. Following the official nuScenes splits~\cite{caesar2020nuscenes}, SpatialLiDAR-QA comprises 89,548 training and 19,263 validation QA pairs, totaling 108,811 pairs across nuScenes keyframes.

\begin{figure*}[t]
\centering
\includegraphics[width=\linewidth]{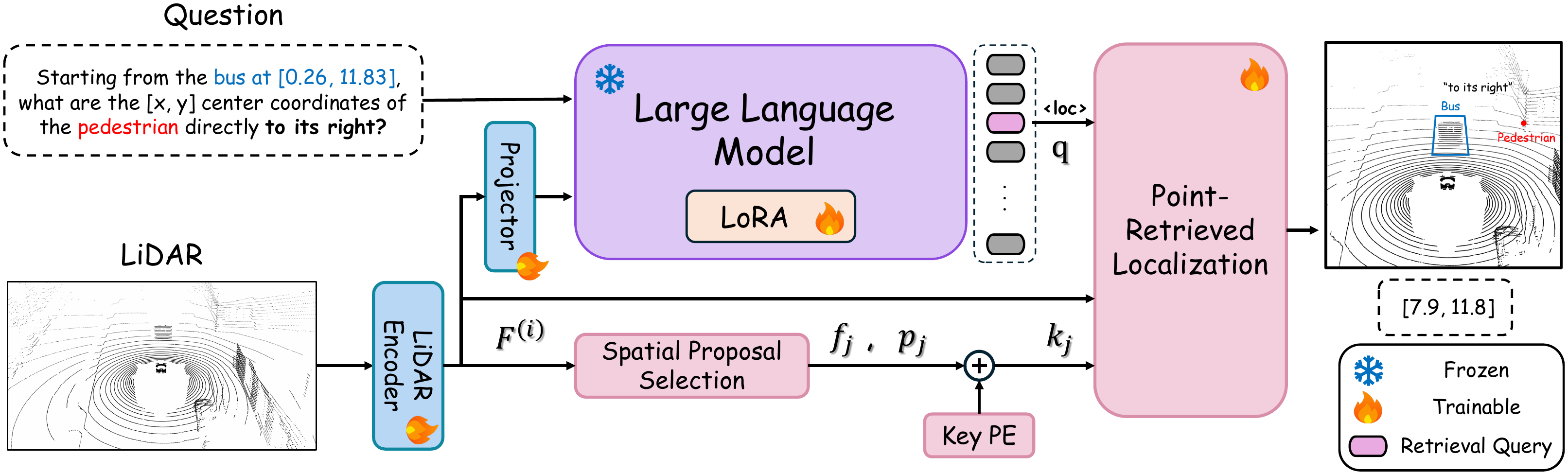}
\caption{Overview of \textbf{SpatialLiDAR-LM}. LiDAR point tokens and
questions are processed in a shared LLM embedding space. For coordinate
prediction, PRL constructs spatial proposals through proposal scoring and
$xy$-NMS, retrieves a question-relevant proposal using the retrieval query
and position-aware proposal keys, and refines its LiDAR-frame coordinate from
local point features.}
\label{fig:overview}
\vspace{-6mm}
\end{figure*}

\subsection{Dataset Construction Process}
\label{subsec:dataset_construction}
SpatialLiDAR-QA is generated on official nuScenes keyframes~\cite{caesar2020nuscenes} using geometry-verified rules across eight object categories (\textit{car, truck, bus, trailer, construction\_vehicle, pedestrian, bicycle, motorcycle}). Strict filtering removes ambiguous directional cases, while maneuver-feasibility labels are balanced to avoid skewed answer distributions. Specifically, following the difficulty criteria of the Waymo Open Dataset~\cite{waymo}, objects containing fewer than 5 LiDAR points are filtered out to eliminate heavily occluded or unidentifiable instances.

We express ground-truth object centers in a BEV frame centered at the LiDAR sensor, with ego-aligned axes such that $+x$ points right and $+y$ forward. Accordingly, \textit{front}, \textit{back}, \textit{right}, and \textit{left} correspond to $+y$, $-y$, $+x$, and $-x$, respectively. Relative displacements are measured from the reference object along these fixed axes rather than its heading, maintaining a consistent directional convention across reference changes in multi-step questions. A directional relation is accepted only when the displacement along the corresponding axis is at least three times the perpendicular displacement magnitude. Thus, diagonal cases remain undefined rather than being assigned a directional label.

For each reference--direction pair, we select an admissible target
within $3$--$20\,\mathrm{m}$, excluding candidates outside this range.
The pair is discarded if the distances of the two nearest candidates
to the reference differ by less than $2\,\mathrm{m}$. We further reject
locally crowded configurations and pairs that fail a geometric-clearance
check along the connecting segment. These filters reduce ambiguity in
target identification when composing relations across multiple steps. Human verification was conducted across QA pairs to manually validate geometric correctness and question--answer consistency.

\section{METHODS}
\label{sec:methods}

We present \textbf{SpatialLiDAR-LM}, an end-to-end architecture that aligns an LLM with a general-purpose 3D point-cloud backbone. SpatialLiDAR-LM preserves point-cloud representations and grounds language in their spatial structure as illustrated in Fig.~\ref{fig:overview}.

\subsection{Model Architecture}
\label{subsec:architecture}

\textbf{LiDAR Point Feature Encoding.}
Rather than using proposals produced by a task-specific 3D detector, we use
PointTransformerV3 (PTv3)~\cite{wu2024pointcept}, initialized from a
checkpoint trained for LiDAR semantic segmentation on
nuScenes~\cite{caesar2020nuscenes}. PTv3 converts the full LiDAR scan into
point-level features, which are projected into the LLM embedding space by
a lightweight two-layer MLP and processed
jointly with the question.

\textbf{Language Model.}
We fine-tune the LLM using LoRA~\cite{hu2022lora} to jointly process the
projected point tokens and the tokenized question, in which coordinates are
discretized into learned bin tokens ($0.25\,\text{m}$ resolution over
$\pm54\,\text{m}$). Coordinate prediction is handled by the PRL module
described in Sec.~\ref{subsec:prl}.

\subsection{Point-Retrieved Localization}
\label{subsec:prl}

Several localization-oriented MLLMs formulate coordinate prediction as sequential token generation, representing locations with textual numbers or discrete location tokens~\cite{chen2023shikra,peng2024kosmos2}. LiDAR-LLM also supports location-token-based grounding~\cite{yang2024lidarllm}. Our contribution is not the introduction of a localization token, but how the localization-token representation is connected to the observed geometry. Rather than using this representation solely to decode coordinates, Point-Retrieved Localization (PRL) uses it to retrieve a question-relevant spatial proposal and derives the final coordinate from the local points supporting that proposal. This provides an explicit geometric readout through proposal retrieval and local point refinement.

\textbf{Spatial Proposal Selection.}
Directly searching over all point tokens retains substantial spatial redundancy. PRL therefore assigns each encoded point feature
$\mathbf{F}^{(i)}$,
corresponding to the $i$-th point token, a query-independent proposal score, \begin{equation} \label{eq:prop_score} a_i = g_{\mathrm{prop}}(\mathbf{F}^{(i)}) + \eta\,\operatorname{sg}\!\left(\|\mathbf{F}^{(i)}\|_2\right), \end{equation} where $g_{\mathrm{prop}}$ is a learned scalar projection and $\operatorname{sg}(\cdot)$ denotes the stop-gradient operator. The feature-energy term ($\eta=0.1$) provides a non-learned ranking signal
before $g_{\mathrm{prop}}$ becomes informative, while detaching it prevents
gradients through this prior from directly encouraging larger feature
magnitudes. The learned component of $a_i$ is optimized through the grounding objective without separate objectness supervision (Sec.~\ref{subsec:training}), and the score does not predict object categories or boxes. PRL then selects up to $M$ spatially separated proposals according to these scores, using non-maximum suppression in the $xy$ plane to avoid repeatedly sampling the same region. Let $\mathbf{o}_j\in\mathbb{R}^2$ denote the initial LiDAR-frame location of the
$j$-th proposal and $\mathbf{c}_i\in\mathbb{R}^2$ the coordinate of the
$i$-th point token. We define the local neighborhood around the $j$-th
proposal as $
\mathcal{N}(j)
=
\{i:\|\mathbf{c}_i-\mathbf{o}_j\|_2 \le r_{\mathrm{agg}}\}.
$
The features and coordinates of the point tokens in $\mathcal{N}(j)$ are
then aggregated as

\begin{equation}
\label{eq:proposal}
\begin{aligned}
w_{ij}
&=
\operatorname{softmax}_{i\in\mathcal{N}(j)}(a_i),\\
\mathbf{f}_j
&=
\sum_{i\in\mathcal{N}(j)}w_{ij}\mathbf{F}^{(i)},
\qquad
\mathbf{p}_j
=
\sum_{i\in\mathcal{N}(j)}w_{ij}\mathbf{c}_i,
\end{aligned}
\end{equation}
where $w_{ij}$ is the normalized weight of the $i$-th point token within
$\mathcal{N}(j)$. As defined in Eq.~\ref{eq:proposal}, each proposal is represented by a local feature $\mathbf{f}_j$ and its LiDAR-frame center $\mathbf{p}_j$.

\textbf{Language-Conditioned Retrieval and Refinement.}
Prior MLLMs extend the language vocabulary with task-specific tokens and use their hidden representations to condition non-linguistic decoders, such as mask prediction in LISA and 3D grounding in LLaVA-3D~\cite{lai2024lisa,zhu2025llava3d}. Following this paradigm, we add a localization token \texttt{<loc>} to the LLM vocabulary and replace each coordinate in the response that requires spatial grounding with this token during training. The language modeling objective therefore teaches the model when a geometric readout is required. Let $\mathbf{h}_{\mathrm{loc}}$ denote the contextual hidden state at the generated \texttt{<loc>} position. PRL projects it into a retrieval query $\mathbf{q}=\mathbf{h}_{\mathrm{loc}}W_q$, which carries the question context and the model's resolved referent. For the $j$-th proposal, we construct a retrieval key $\mathbf{k}_j$ using
its aggregated feature $\mathbf{f}_j$ and positional information, as
described in Sec.~\ref{subsec:key_pe}.

\begin{figure}[t]
  \centering
  \includegraphics[width=\columnwidth]{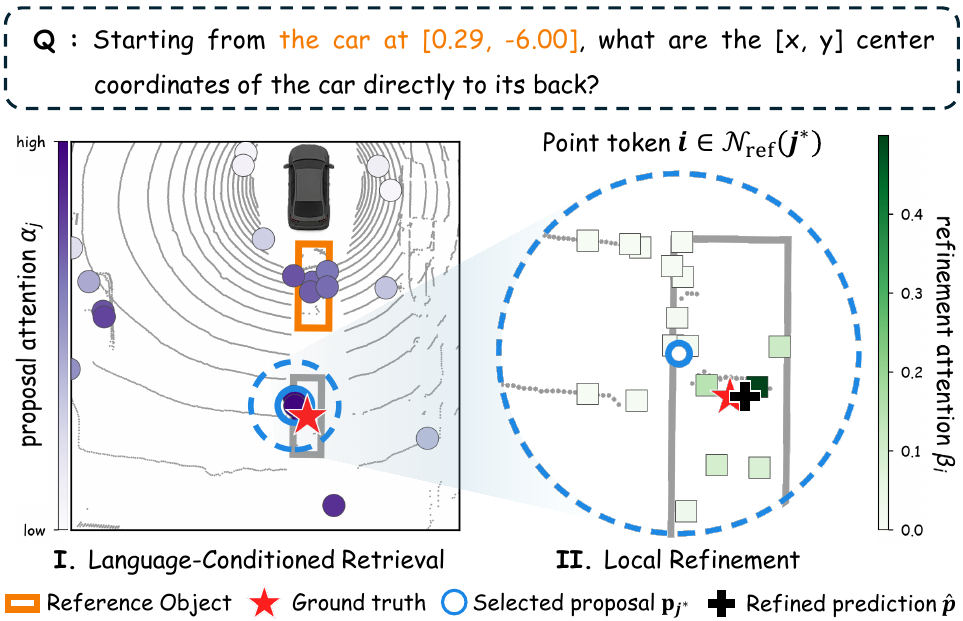}
  \caption{Point-Retrieved Localization (PRL) on a validation example.
  The retrieval query first selects a question-relevant spatial proposal
  using proposal-level attention $\alpha_j$, after which local attention
  $\beta_i$ over the point tokens in $\mathcal{N}_{\mathrm{ref}}$ refines the final
  coordinate. The red box denotes the ground-truth bounding box.}
  \label{fig:qualitative}
\vspace{-4mm}
\end{figure}

The query is matched against the proposal keys using scaled dot-product attention to retrieve the relevant scene region:
\begin{equation}
\label{eq:prl_retrieval}
\begin{aligned}
s_j
&=
\frac{\mathbf{q}\mathbf{k}_j^\top}{\sqrt{d_{\mathrm{attn}}}},
\qquad
\alpha_j
=
\operatorname{softmax}(\mathbf{s})_j,\\
\tilde{\mathbf{p}}
&=
\sum_{j=1}^{M}\alpha_j\mathbf{p}_j,
\qquad
j^*
=
\arg\max_j s_j.
\end{aligned}
\end{equation}
As shown in Eq.~\ref{eq:prl_retrieval}, the soft proposal distribution yields the coarse coordinate $\tilde{\mathbf{p}}$, while $j^*$ selects the local region used for refinement. PRL then compares the query with the point tokens around the selected center, $\mathcal{N}_{\mathrm{ref}}(j^*)
=
\left\{
i :
\|\mathbf{c}_i-\mathbf{p}_{j^*}\|_2
\le r_{\mathrm{agg}}
\right\}$, and predicts the final coordinate as
\begin{equation}
\label{eq:prl_refine}
\hat{\mathbf{p}}
=
\sum_{i\in\mathcal{N}_{\mathrm{ref}}(j^*)}\beta_i\mathbf{c}_i,
\qquad
\beta_i
=
\operatorname{softmax}\left(
\frac{\mathbf{q}(\mathbf{F}^{(i)}W_k')^\top}
{\sqrt{d_{\mathrm{attn}}}}
\right)_i.
\end{equation}
Here, $\alpha_j$ and $\beta_i$ denote the proposal-level and local point-level attention weights, respectively. The refinement in Eq.~\ref{eq:prl_refine} grounds the final prediction in the observed geometry by taking an attention-weighted combination of the local point coordinates. When the selected proposal has insufficient point support, PRL retains
the coarse estimate $\tilde{\mathbf{p}}$. Fig.~\ref{fig:qualitative} shows both stages on a validation example. 

\subsection{Proposal-Key Positional Encoding (Key PE)}
\label{subsec:key_pe}

\textbf{Spatial Ambiguity.}
Proposal features describe the local geometry around each candidate, but they may not sufficiently distinguish multiple instances with similar structure. This is common in driving scenes containing several vehicles or pedestrians, where semantic similarity alone can lead the query to retrieve the wrong instance. The issue becomes more pronounced for relational questions, which require the model to distinguish candidates using both their semantic content and their LiDAR-frame position.

\textbf{Position-Aware Proposal Keys.}
To expose this spatial information during retrieval, we augment each semantic proposal key with an encoding of its LiDAR-frame center:
\begin{equation}
\label{eq:keype}
\mathbf{k}_j
=
\mathbf{f}_jW_k
+
W_p\phi(\mathbf{p}_j).
\end{equation}
As shown in Eq.~\ref{eq:keype}, the positional term is added only to the proposal key used for retrieval. For a LiDAR-frame proposal center normalized by the perceptual range $R$, we use Fourier features~\cite{tancik2020fourier} over $B$ logarithmically spaced frequency bands:
\begin{equation}
\label{eq:fourier}
\phi(\mathbf{p}_j)
=
\left[
\sin\left(\frac{\pi 2^b\mathbf{p}_j}{R}\right),
\cos\left(\frac{\pi 2^b\mathbf{p}_j}{R}\right)
\right]_{b=0}^{B-1}
\in\mathbb{R}^{4B}.
\end{equation}
Equation~\ref{eq:fourier} exposes the proposal location at multiple spatial scales, allowing the query to distinguish geometrically similar candidates by position while retaining their semantic features. The positional projection $W_p$ is initialized to zero, so the model starts from the position-agnostic proposal keys and gradually learns the positional residual.

\begin{table*}[t!]
\centering
\caption{Quantitative results (accuracy, \%, within $0.5$, $1.0$, and
$2.0\,\text{m}$) on Coordinate prediction tasks. Language CE predicts
coordinates through language decoding instead of PRL; best results per
base LLM are in \textbf{bold}.}
\label{tab:coord_results}

\resizebox{\textwidth}{!}{%
\setlength{\tabcolsep}{4pt}
\begin{tabular}{l l c rrr rrr rrr}
\toprule
\textbf{Base LLM} & \textbf{Method} & \textbf{Modality}
& \multicolumn{3}{c}{\textbf{Localization}}
& \multicolumn{3}{c}{\textbf{Single-Step}}
& \multicolumn{3}{c}{\textbf{Multi-Step}} \\
\cmidrule(lr){4-6}
\cmidrule(lr){7-9}
\cmidrule(lr){10-12}
& & &
@0.5m & @1.0m & @2.0m &
@0.5m & @1.0m & @2.0m &
@0.5m & @1.0m & @2.0m \\
\midrule

\multirow[c]{6}{*}{SmolVLM}
& SmolVLM-500M-Instruct
& Multi-Camera
& 14.9 & 30.5 & 52.3
& 8.2 & 24.5 & 51.5
& 3.9 & 11.7 & 26.8 \\

& BEV-LLM~\cite{11097781}
& Multi-Camera
& 2.4 & 4.1 & 6.5
& 8.5 & 23.5 & 49.1
& 3.8 & 11.2 & 25.3 \\

& BEV-LLM~\cite{11097781}
& LiDAR
& 2.6 & 4.3 & 6.5
& 8.1 & 24.0 & 49.2
& 3.4 & 12.2 & 24.1 \\

& LiDAR-LLM~\cite{yang2024lidarllm}
& LiDAR
& 0.5 & 1.6 & 5.3
& 3.1 & 11.5 & 34.7
& 0.9 & 5.8 & 21.8 \\

& \textit{SpatialLiDAR-LM (Language CE)}
& LiDAR
& 6.7 & 15.6 & 31.9
& 3.6 & 9.3 & 22.4
& 2.2 & 8.7 & 20.0 \\

\cmidrule(lr){2-12}

& \textbf{SpatialLiDAR-LM (Ours)}
& LiDAR
& \textbf{57.2} & \textbf{77.0} & \textbf{85.0}
& \textbf{28.4} & \textbf{46.3} & \textbf{59.9}
& \textbf{36.4} & \textbf{53.3} & \textbf{63.0} \\

\midrule

\multirow[c]{6}{*}{Qwen3.5}
& Qwen3.5-0.8B
& Multi-Camera
& 38.5 & 63.1 & 79.3
& 12.4 & 32.5 & 58.9
& 8.5 & 21.5 & 42.4 \\

& BEV-LLM~\cite{11097781}
& Multi-Camera
& 4.5 & 6.8 & 12.0
& 9.4 & 25.7 & 51.8
& 5.3 & 13.4 & 27.6 \\

& BEV-LLM~\cite{11097781}
& LiDAR
& 2.4 & 4.3 & 6.4
& 10.6 & 26.0 & 49.3
& 3.8 & 12.6 & 25.6 \\

& LiDAR-LLM~\cite{yang2024lidarllm}
& LiDAR
& 0.6 & 3.6 & 6.7
& 5.8 & 16.3 & 42.6
& 1.6 & 4.8 & 18.4 \\

& \textit{SpatialLiDAR-LM (Language CE)}
& LiDAR
& 10.3 & 22.7 & 41.0
& 6.4 & 19.0 & 45.5
& 5.8 & 12.7 & 29.5 \\

\cmidrule(lr){2-12}

& \textbf{SpatialLiDAR-LM (Ours)}
& LiDAR
& \textbf{57.6} & \textbf{78.2} & \textbf{85.4}
& \textbf{34.7} & \textbf{56.2} & \textbf{71.4}
& \textbf{44.1} & \textbf{62.9} & \textbf{74.0} \\

\bottomrule
\end{tabular}%
}
\vspace{-4mm}
\end{table*}

\subsection{Training Objective and Inference}
\label{subsec:training}

SpatialLiDAR-LM is trained jointly for language generation and coordinate prediction using the Smooth L1 loss:
\begin{equation}
\label{eq:loss_prl}
\mathcal{L}_{\text{total}}
=
\mathcal{L}_{\text{LM}}
+
\lambda_{\text{PRL}}
\left[
\operatorname{SmoothL1}(\hat{\mathbf{p}},\mathbf{p}_{\text{gt}})
+
\lambda_{\text{con}}\mathcal{L}_{\text{con}}
\right].
\end{equation}
Following the multi-positive formulation of supervised contrastive learning~\cite{khosla2020supervised}, the instance-aware contrastive term is defined over the proposal retrieval scores from Eq.~\ref{eq:prl_retrieval}:
\begin{equation}
\label{eq:loss_con}
\mathcal{L}_{\text{con}}
=
-\frac{1}{|\mathcal{P}^{+}|}
\sum_{j\in\mathcal{P}^{+}}
\log
\frac{\exp(s_j/\tau)}
{\sum_{m=1}^{M}\exp(s_m/\tau)},
\end{equation}
where $\mathcal{P}^{+}$ contains proposals whose centers fall within the target box enlarged by $0.5\,\text{m}$ along each horizontal axis; when the box extent is unavailable, proposals within $2.0\,\text{m}$ of $\mathbf{p}_{\text{gt}}$ are used. The contrastive term is omitted when no proposal satisfies this criterion, and $\tau$ is a fixed temperature. In Eq.~\ref{eq:loss_prl}, $\mathcal{L}_{\text{LM}}$ is the causal language modeling loss, the Smooth L1 term supervises point-level coordinate refinement, and Eq.~\ref{eq:loss_con} directly trains the language-conditioned score $s_j$ to retrieve target-consistent proposals. The query-independent score $a_i$ instead selects the proposal set. Seed selection and NMS are discrete, but the local aggregation in Eq.~\ref{eq:proposal} remains differentiable, allowing $\mathcal{L}_{\text{con}}$ to update $g_{\mathrm{prop}}$ through the resulting proposal features.

At inference, PRL constructs spatial proposals directly from the encoded point tokens and requires neither target annotations nor an external detector.

\section{EXPERIMENTS}
\label{sec:experiments}

\subsection{Experimental Setup}
\label{subsec:setup}

\textbf{Implementation Details.} 
The LiDAR encoder is PointTransformerV3 (PTv3)~\cite{wu2024pointcept}, pretrained on nuScenes, following its original preprocessing setup. As base LLMs, we use Qwen3.5-0.8B~\cite{qwen2026qwen35} and SmolVLM-500M-Instruct~\cite{marafioti2025smolvlm}, with their vision encoders removed. For PRL, we empirically select $M=96$ after evaluating $M$ ranging from 48 to 128. We set $r_{\mathrm{NMS}}=2.0\,\text{m}$ and $r_{\mathrm{agg}}=2.5\,\text{m}$. Key PE uses $B=6$ frequency bands with perceptual range $R=54\,\text{m}$.

The model is optimized end-to-end using AdamW~\cite{loshchilov2019decoupled} with $\beta_1=0.9$, $\beta_2=0.95$, and weight decay $0.01$. LoRA ($r=64$, $\alpha=128$, dropout 0.05) is applied to all attention projection layers. We use cosine annealing with a warmup ratio of $0.03$ and peak learning rates of $2\times10^{-4}$ for the LoRA adapters and point projector and $1\times10^{-5}$ for PTv3. The loss weights are $\lambda_{\mathrm{PRL}}=\lambda_{\mathrm{con}}=0.5$, with temperature $\tau=0.07$. Training and evaluation use NVIDIA RTX A6000 GPUs ($48\,\text{GB}$ VRAM each); training runs for 2 epochs on four GPUs with a per-device batch size of 4 (effective batch size 16).

\textbf{Baselines and Evaluation Metrics.}
We compare SpatialLiDAR-LM against LiDAR-LLM~\cite{yang2024lidarllm},
BEV-LLM~\cite{11097781} with multi-camera and LiDAR inputs, and
multi-camera VLM baselines.
All baselines are re-implemented with the same base LLMs and trained on
SpatialLiDAR-QA with the same prompt templates and LoRA schedule as our
model. Following the multi-view setup used in driving VLMs and VLAs~\cite{ding2024bevinmllm,tian2025drivevlm,zhou2025autovla}, the VLM baselines take synchronized surround-view images as input. For
Localization and coordinate prediction tasks, we report accuracy within $0.5\,\text{m}$,
$1.0\,\text{m}$, and $2.0\,\text{m}$ of the ground-truth location. For
Class prediction and Maneuver Feasibility, we report exact-match accuracy. Unless otherwise specified, results are reported on the SpatialLiDAR-QA validation split.

\begin{table}[h]
\centering
\caption{Ablation of the grounding mechanism and Key PE.
Accuracy (\%) is reported within $0.5$, $1.0$, and
$2.0\,\text{m}$ of the ground-truth location.}
\label{tab:ablation_results}
\resizebox{\columnwidth}{!}{%
\setlength{\tabcolsep}{4pt}
\begin{tabular}{l l c cccc}
\toprule
\textbf{Base LLM} & \textbf{Task} & \textbf{Threshold}
& \textbf{Language CE} & \textbf{Smooth L1}
& \textbf{PRL} & \textbf{PRL + Key PE} \\
\midrule

\textbf{SmolVLM}
& \textbf{Localization} & @0.5m & 6.7 & 1.0 & 54.8 & \textbf{57.2} \\
& & @1.0m & 15.6 & 4.1 & 76.5 & \textbf{77.0} \\
& & @2.0m & 31.9 & 14.7 & 84.6 & \textbf{85.0} \\
\cmidrule(lr){2-7}

& \textbf{Single-Step} & @0.5m & 3.6 & 5.9 & 14.5 & \textbf{28.4} \\
& & @1.0m & 9.3 & 16.9 & 25.7 & \textbf{46.3} \\
& & @2.0m & 22.4 & 42.8 & 38.7 & \textbf{59.9} \\
\cmidrule(lr){2-7}

& \textbf{Multi-Step} & @0.5m & 2.2 & 1.2 & 17.9 & \textbf{36.4} \\
& & @1.0m & 8.7 & 4.9 & 27.8 & \textbf{53.3} \\
& & @2.0m & 20.0 & 20.9 & 35.8 & \textbf{63.0} \\

\midrule

\textbf{Qwen3.5}
& \textbf{Localization} & @0.5m & 10.3 & 3.2 & 53.4 & \textbf{57.6} \\
& & @1.0m & 22.7 & 11.5 & 74.4 & \textbf{78.2} \\
& & @2.0m & 41.0 & 29.9 & 82.3 & \textbf{85.4} \\
\cmidrule(lr){2-7}

& \textbf{Single-Step} & @0.5m & 6.4 & 5.2 & 26.2 & \textbf{34.7} \\
& & @1.0m & 19.0 & 18.7 & 44.1 & \textbf{56.2} \\
& & @2.0m & 45.5 & 46.3 & 56.2 & \textbf{71.4} \\
\cmidrule(lr){2-7}

& \textbf{Multi-Step} & @0.5m & 5.8 & 1.2 & 32.8 & \textbf{44.1} \\
& & @1.0m & 12.7 & 4.8 & 51.1 & \textbf{62.9} \\
& & @2.0m & 29.5 & 19.6 & 60.3 & \textbf{74.0} \\

\bottomrule
\end{tabular}%
}
\vspace{-4mm}
\end{table}

\subsection{Quantitative Results}
\label{subsec:quantitative_results}

\textbf{Coordinate prediction tasks.}
In Table~\ref{tab:coord_results}, SpatialLiDAR-LM achieves the highest accuracy on Localization and Single/Multi-Step Relational Grounding across all distance thresholds for both base LLMs. With Qwen3.5 at $0.5\,\text{m}$, SpatialLiDAR-LM reaches 57.6\% on Localization, compared with 38.5\% for the multi-camera VLM, and 44.1\% on Multi-Step Relational Grounding, substantially higher than the VLM's 8.5\%. With both SmolVLM and Qwen3.5, the same proposal retrieval and local point refinement design yields consistent gains in localization and relational grounding. This consistency shows that SpatialLiDAR-LM is compatible with different base LLMs.

While other models struggle on relational grounding, SpatialLiDAR-LM achieves substantially higher accuracy on both Single-Step and Multi-Step, suggesting that it composes spatial relations reliably. Its lower accuracy on Single-Step than on Multi-Step is partly due to occlusion: since a Single-Step target lies close to its reference object, it is more often partially occluded and captured by fewer LiDAR points.

\textbf{Grounding Mechanism Analysis.}
We further examine the grounding mechanism by comparing alternative
coordinate prediction formulations using the same point encoder and
base LLM (Table~\ref{tab:ablation_results}). Language CE generates
coordinates as text using token-level cross-entropy. The Smooth L1
baseline retains the same \texttt{<loc>} token as PRL but directly
regresses coordinates from its contextual hidden state using the same
Smooth L1 coordinate loss. With Qwen3.5, Localization accuracy at $0.5\,\text{m}$ is 10.3\% for Language CE and 3.2\% for Smooth L1, compared with 53.4\% for PRL.

At the same $0.5\,\text{m}$ threshold, PRL also increases Localization
accuracy from 6.7\% to 54.8\% with SmolVLM. On relational grounding,
the gains over Language CE are 10.9 and 19.8 percentage points for
Single-Step and 15.7 and 27.0 points for Multi-Step with SmolVLM and
Qwen3.5, respectively. Smooth L1 also remains below PRL on all three
tasks with both base LLMs at this threshold.
These results highlight the benefit of using the contextual
representation to retrieve question-relevant proposals and derive
coordinates from their supporting local LiDAR points, rather than
directly decoding or regressing coordinates.

\textbf{Key PE Analysis.}
Key PE provides only modest gains on Localization but substantially larger
gains on relational grounding. At $0.5\,\text{m}$, Localization accuracy
improves by 2.4 percentage points with SmolVLM and 4.2 points with
Qwen3.5, whereas Single-Step improves by 13.9 and 8.5 points,
respectively, and Multi-Step by 18.5 and 11.3 points. This indicates that
proposal-level positional information is particularly important for
retrieving the correct proposal under relational constraints. Both the
gains from PRL and the larger relational gains from Key PE are consistent
across the two base LLMs.

\begin{table}[h]
\centering
\caption{Quantitative results (exact-match accuracy, \%) on Class prediction tasks and Maneuver Feasibility. C and L denote multi-camera and LiDAR.}
\label{tab:semantic_results}
\resizebox{\columnwidth}{!}{%
\setlength{\tabcolsep}{3pt}
\begin{tabular}{c l c cccc}
\toprule
\textbf{Base LLM} & \textbf{Method} & \textbf{Modality}
& \textbf{Maneuver} & \textbf{Object Identification}
& \textbf{Single-Step} & \textbf{Multi-Step} \\
\midrule

\multirow[c]{5}{*}{SmolVLM}
& SmolVLM-500M-Instruct
& C
& \textbf{89.7} & \textbf{62.2} & \textbf{64.0} & 71.5 \\

& BEV-LLM~\cite{11097781}
& C
& 60.6 & 24.0 & 58.8 & 70.4 \\

& BEV-LLM~\cite{11097781}
& L
& 60.7 & 26.0 & 59.0 & 70.8 \\

& LiDAR-LLM~\cite{yang2024lidarllm}
& L
& 69.0 & 16.0 & 57.0 & 64.6 \\

& \textbf{SpatialLiDAR-LM (Ours)}
& L
& 87.2 & 38.0 & 59.4 & \textbf{71.8} \\

\midrule

\multirow[c]{5}{*}{Qwen3.5}
& Qwen3.5-0.8B
& C
& \textbf{92.2} & \textbf{77.1} & \textbf{75.9} & \textbf{80.0} \\

& BEV-LLM~\cite{11097781}
& C
& 81.9 & 29.5 & 60.2 & 72.7 \\

& BEV-LLM~\cite{11097781}
& L
& 60.5 & 28.0 & 60.1 & 72.7 \\

& LiDAR-LLM~\cite{yang2024lidarllm}
& L
& 59.4& 27.1& 60.3& 71.6\\

& \textbf{SpatialLiDAR-LM (Ours)}
& L
& 85.5 & 73.0 & 58.2 & 72.4 \\

\bottomrule
\end{tabular}%
}
\vspace{-4mm}
\end{table}

\textbf{Class prediction tasks and Maneuver Feasibility.}
Table~\ref{tab:semantic_results} evaluates categorical text outputs: Object Identification, the class predictions of Single- and Multi-Step Relational Grounding, and Maneuver Feasibility. The VLMs generally achieve the highest accuracy on these tasks, consistent with the richer semantic and contextual cues available in images. For example, with Qwen3.5, the multi-camera VLM achieves 92.2\% on Maneuver Feasibility, compared with 85.5\% for SpatialLiDAR-LM. Nevertheless, SpatialLiDAR-LM achieves higher or competitive performance compared with the other LiDAR-based baselines on most tasks. In particular, on Maneuver Feasibility, it reaches 87.2\% with SmolVLM and 85.5\% with Qwen3.5, substantially outperforming the LiDAR-based BEV-LLM and LiDAR-LLM baselines.
SpatialLiDAR-LM is also competitive with the multi-camera VLM on selected
semantic tasks. With SmolVLM, it achieves 71.8\% on Multi-Step class
prediction, compared with 71.5\% for the VLM; with Qwen3.5, it reaches
73.0\% on Object Identification, compared with 77.1\% for the VLM.
These results indicate that the proposed LiDAR--language architecture
supports semantic recognition alongside precise coordinate grounding.


\begin{table}[h]
\centering
\caption{Cross-dataset grounding accuracy (\%).
Both models use Qwen3.5-0.8B and are fine-tuned on
nuScenes-based SpatialLiDAR-QA; evaluation on AV2 and NAVSIM
is zero-shot. Distance thresholds are in meters.}
\label{tab:cross_dataset}

\resizebox{\columnwidth}{!}{%
\begin{tabular}{@{}llccc ccc ccc@{}}
\toprule
\multirow{2}{*}{Eval. dataset}
& \multirow{2}{*}{Model}
& \multicolumn{3}{c}{Localization}
& \multicolumn{3}{c}{Single-Step}
& \multicolumn{3}{c}{Multi-Step} \\
\cmidrule(lr){3-5} \cmidrule(lr){6-8} \cmidrule(lr){9-11}
& & @0.5 & @1.0 & @2.0
  & @0.5 & @1.0 & @2.0
  & @0.5 & @1.0 & @2.0 \\
\midrule

\multirow{2}{*}{\textcolor{gray}{nuScenes (source)}}
& \textcolor{gray}{Ours}
& \textcolor{gray}{\textbf{57.6}} & \textcolor{gray}{\textbf{78.2}} & \textcolor{gray}{\textbf{85.4}}
& \textcolor{gray}{\textbf{34.7}} & \textcolor{gray}{\textbf{56.2}} & \textcolor{gray}{\textbf{71.4}}
& \textcolor{gray}{\textbf{44.1}} & \textcolor{gray}{\textbf{62.9}} & \textcolor{gray}{\textbf{74.0}} \\
& \textcolor{gray}{VLM}
& \textcolor{gray}{38.5} & \textcolor{gray}{63.1} & \textcolor{gray}{79.3}
& \textcolor{gray}{12.4} & \textcolor{gray}{32.5} & \textcolor{gray}{58.9}
& \textcolor{gray}{8.5} & \textcolor{gray}{21.5} & \textcolor{gray}{42.4} \\
\midrule

\multirow{2}{*}{AV2~\cite{Argoverse2}}
& Ours
& \textbf{47.1} & \textbf{67.8} & \textbf{76.2}
& \textbf{36.3} & \textbf{55.4} & \textbf{65.3}
& \textbf{31.6} & \textbf{50.4} & \textbf{57.9} \\
& VLM
& 12.1 & 28.9 & 54.6
& 7.6 & 20.7 & 44.2
& 1.7 & 6.7 & 23.6 \\
\midrule

\multirow{2}{*}{NAVSIM~\cite{Dauner2024NEURIPS}}
& Ours
& \textbf{67.8} & \textbf{77.3} & \textbf{79.9}
& \textbf{43.5} & \textbf{57.4} & \textbf{63.2}
& \textbf{41.4} & \textbf{54.4} & \textbf{59.5} \\
& VLM
& 6.3 & 18.0 & 34.9
& 9.6 & 25.3 & 49.0
& 3.3 & 10.8 & 29.2 \\
\bottomrule
\end{tabular}%
}
\vspace{-4mm}
\end{table}

\subsection{Cross-Dataset Evaluation and Efficiency}
We compare the zero-shot cross-dataset performance of SpatialLiDAR-LM
and the multi-camera VLM on the Argoverse 2
(AV2)~\cite{Argoverse2} validation split and the
NAVSIM~\cite{Dauner2024NEURIPS} test split. QA pairs for both datasets are generated using the same pipeline as
SpatialLiDAR-QA (Sec.~\ref{subsec:dataset_construction}). Both models use
Qwen3.5-0.8B and are fine-tuned on the same nuScenes-based
SpatialLiDAR-QA training split, without target-dataset fine-tuning.
We use the coordinate-accuracy metrics defined in
Sec.~\ref{subsec:setup} and map target-dataset object classes to our
taxonomy: AV2 covers seven of our eight classes, while NAVSIM
provides vehicle, pedestrian, and bicycle labels. The VLM uses
surround-view images as input.

As shown in Table~\ref{tab:cross_dataset}, SpatialLiDAR-LM
outperforms the multi-camera VLM on all three coordinate prediction
tasks at every distance threshold on both target datasets.
At $0.5\,\text{m}$, Localization accuracy reaches 47.1\% on AV2 and
67.8\% on NAVSIM, compared with 12.1\% and 6.3\% for the VLM,
respectively. The consistent advantage across direct localization
and relational grounding supports stronger zero-shot cross-dataset
generalization of SpatialLiDAR-LM under the same source-dataset
fine-tuning setup. These results suggest that language-conditioned
proposal retrieval and local geometric readout transfer effectively
beyond nuScenes.

We measure the per-query inference latency of SpatialLiDAR-LM over all
19,263 validation queries on an NVIDIA RTX A6000 with batch size 1.
With Qwen3.5-0.8B as the language backbone, SpatialLiDAR-LM has a mean
latency of $574.4\,\mathrm{ms}$ and peak allocated GPU memory of
$5.60\,\mathrm{GB}$. With SmolVLM-500M-Instruct as the language backbone,
SpatialLiDAR-LM has a mean latency of $490.9\,\mathrm{ms}$ and peak
allocated GPU memory of $4.76\,\mathrm{GB}$.

\section{CONCLUSION}
\label{sec:conclusion}

In this work, we advance LiDAR--language understanding by connecting
language-based spatial reasoning with the metric geometry of outdoor
scenes. We introduce SpatialLiDAR-QA to support training and evaluation
on tasks that require identifying objects, following or composing spatial
relations, and grounding the intended target. SpatialLiDAR-LM addresses
these tasks through language-conditioned retrieval and local point-based
coordinate prediction, linking the LLM's interpretation of a question
to the observed LiDAR geometry. Experiments and ablation studies support the proposed design for precise
localization and relational grounding. Competitive performance on selected
semantic tasks and zero-shot cross-dataset results further suggest that
this approach can support broader LiDAR--language understanding.
Together, SpatialLiDAR-QA and SpatialLiDAR-LM provide a foundation for
developing geometrically grounded language models that integrate semantic
recognition, spatial reasoning, and precise localization. We will publicly
release the dataset and training code to support further research toward
language-guided spatial understanding in autonomous driving and outdoor
robotics.

\textbf{Limitations.}
While our approach effectively aligns language
with precise LiDAR coordinates, the current formulation focuses on
explicit directional relations and question-relevant localization
within a shared spatial frame. Future work will focus on strengthening
semantic understanding and extending spatial grounding to temporal
reasoning to support a broader range of tasks in autonomous driving
and outdoor robotics.


\bibliographystyle{IEEEtran}
\bibliography{references}

\end{document}